%% file: iclr2027_conference.tex
\RequirePackage[svgnames,table]{xcolor}
\documentclass[11pt,letterpaper]{mystyle}
\usepackage[all]{hypcap}
\usepackage[numbers,sort&compress]{natbib}
\usepackage{hyperref}
\hypersetup{colorlinks=true, citecolor={DodgerBlue}, linkcolor={DodgerBlue}, urlcolor={DodgerBlue}}
\usepackage{multicol}
\usepackage{setspace}
\usepackage{caption}
\usepackage{ragged2e}
\usepackage{xcolor}
\usepackage[normalem]{ulem}

\usepackage[most]{tcolorbox}
\tcbuselibrary{skins,breakable}
\usepackage{url}
\usepackage{subcaption}   % subcaption-style \begin{subfigure}{width} panels
\usepackage{listings}
\usepackage{titlesec}
\usepackage{booktabs}
\usepackage{multicol}
\usepackage{multirow}
\usepackage{makecell}
\usepackage{cuted}
\usepackage{array}       % For custom column types (e.g., p{3.3cm})
\usepackage{caption}     % For better table captions
\usepackage{dsfont}
\usepackage{wrapfig}
\usepackage{graphicx}
\usepackage{tabularx}
\usepackage{cleveref}

\usepackage{algorithm}
\usepackage{algpseudocode}
\usepackage{amsmath}
\usepackage{amssymb}
\usepackage{enumitem}
\newenvironment{itemize*}%
 {\leftmargini=20pt\begin{itemize}%
  \setlength{\itemsep}{3pt}%
  \setlength{\parskip}{0pt}%
  }%
 {\end{itemize}} 
\newenvironment{enumerate*}%
 {\begin{enumerate}%
  \setlength{\itemsep}{0pt}%
  \setlength{\parskip}{0pt}}%
 {\end{enumerate}}

\definecolor{teacherblue}{RGB}{35,77,120}
\definecolor{studentorange}{RGB}{158,84,31}
\lstdefinestyle{casecode}{
  language=Python,
  basicstyle=\ttfamily\footnotesize,
  keywordstyle=\color{teacherblue},
  commentstyle=\color{black!55},
  stringstyle=\color{black},
  columns=fullflexible,
  keepspaces=true,
  showstringspaces=false,
  breaklines=true,
  breakatwhitespace=false,
  breakindent=1em,
  aboveskip=5pt,
  belowskip=5pt
}
\newtcolorbox{casestudy}[1]{
  enhanced,
  breakable,
  colback=blue!2,
  colframe=teacherblue,
  boxrule=0.6pt,
  arc=1mm,
  left=1.5mm,
  right=1.5mm,
  top=1mm,
  bottom=1mm,
  before skip=6pt,
  after skip=8pt,
  fonttitle=\bfseries,
  fontupper=\small,
  title={#1}
}
\definecolor{builderblue}{RGB}{35,77,120}
\definecolor{targetgreen}{RGB}{34,117,67}
\definecolor{metaskillwine}{RGB}{128,36,62}
\DeclareRobustCommand{\Builder}[1][Builder]{\textcolor{builderblue}{\textbf{#1}}}
\DeclareRobustCommand{\Target}[1][Target]{\textcolor{targetgreen}{\textbf{#1}}}
\DeclareRobustCommand{\MetaSkill}[1][meta-skill]{\textcolor{metaskillwine}{\textbf{#1}}}

\newcommand{\B}{\mathcal{B}}
\newcommand{\T}{\mathcal{T}}
\newcommand{\D}{\mathcal{D}}

\NewDocumentCommand{\heng}
{ mO{} }{\textcolor{red}{\textsuperscript{\textit{Heng}}\textsf{\textbf{\small[#1]}}}}

\NewDocumentCommand{\zhenhailong}
{ mO{} }{\textcolor{blue}{\textsuperscript{\textit{Zhenhailong}}\textsf{\textbf{\small[#1]}}}}

\title{Learning Meta-Skills for Agent Harness Design in Test-Time AI4AI}

\author{Cheng Qian, Kunlun Zhu, Beibin Li, Zhenhailong Wang, Heng Ji}

\runningtitle{MetaSkill for Test-Time AI4AI}

\begin{document}

\input{sections/0_abstract}
\maketitle

\begingroup
\makeatletter
\renewcommand{\thefootnote}{}
\renewcommand{\@makefntext}[1]{%
  \noindent
  \makebox[0pt][r]{\textsuperscript{$\dagger$}\hspace{0.4em}}%
  #1%
}
\footnotetext{We gratefully acknowledge Mr. Tianqiao Chen, the project lead, for guidance and support throughout this work. The project code is released at \url{https://github.com/qiancheng-apodex/MetaSkill-AI4AI}.
}
\makeatother
\endgroup

\input{sections/1_introduction}

\input{sections/2_related_works}
\input{sections/3_preliminary}
\input{sections/4_method}
\input{sections/5_experiment}
\input{sections/6_analysis}
\input{sections/7_conclusion}

\bibliography{iclr2027_conference}
\bibliographystyle{iclr2027_conference}

\input{sections/Appendix}

\end{document}

%% file: sections/0_abstract.tex
\begin{abstract}
Agent performance depends on both reasoning ability and the environment in which it acts.
We study test-time AI-for-AI, asking how a \emph{\Builder{}} can learn to construct better execution environments for a \emph{\Target{}} while both models’ weights remain fixed.
To make the Builder's experience reusable, we introduce \emph{\MetaSkill[meta-skills]{}}: principles specifying when support is needed and what resources to provide. The Builder learns these principles from Target's execution feedback on the development set, then uses the frozen skill bank to construct harnesses for unseen tasks. Across Harness-Bench and NewtonBench, full-bank meta-skills improve macro-average performance by 8.95 percentage points over no-skill construction, and 12.02 points over direct delivery of the same bank to the Target.
These results highlight the value of translating experience into executable support. Gains when the same model serves both roles further suggest a path to system level self-improvement through learning to build better environments.
\vspace{6mm}
\end{abstract}
% \zhenhailong{I think we need to emphasize more on (1) why this clear separation is necessary/helpful, compared to prior work to mixed these roles like metaRSI; (2) we take the unique perspective of focusing on the Builder, which does not directly relates to optimizing the task performance; (I think we can even add a small teaser figure on this)}
% \zhenhailong{for the benchmark, if we set this paper in an AI-for-AI setting, I think this might trigger reviewers questions: for example why we did not include the popular ones like AI4AI-Bench, PostTrainBench, MLE-Bench etc that has a stronger AI-training-AI flavor. one potential way to avoid this is that we can formulate this as an broader RSI setting (a system self-improve on any tasks) but emphasize that it is important to use this clearer Builder-Target view to separate the optimization landscape. As in our intro we also noted that Builder and Target can be same backbone model.}

%% file: sections/1_introduction.tex
\section{Introduction}
An AI agent's performance depends on both its reasoning ability and the environment in which it acts~\citep{yang2024sweagent}. Consider a capable PhD student who spends days repeating experiments because configurations and results are scattered across scripts and notes. An effective advisor can address this bottleneck by establishing a shared experiment log, reproducible tools, and a clear validation workflow, helping the student devote more effort to scientific judgment. This analogy suggests a complementary direction for improving AI agents: \textbf{learning how to provide the support that makes their existing capabilities more effective}~\citep{lee2026metaharness, ye2026meta}.

\textbf{AI-for-AI} (AI4AI) at test-time offers a route to this goal by enabling AI systems to design agent programs, workflows, and harnesses~\citep{hu2025automated,zhang2025aflow,lee2026metaharness}. This form of AI4AI involves two complementary roles: a \Builder{}, which designs and provides support, and a \Target{}, which uses that support to solve tasks. We study this relationship through harness construction, where the Builder creates a \emph{harness} comprising instructions, resources, and executable mechanisms for the Target. Just as an advisor learns from a PhD student's progress to refine their guidance, the Builder can learn from the Target's execution to improve its support. Building on recent work on meta-level learning~\citep{ye2026meta}, we therefore ask: \textbf{how can the Builder turn the outcomes of its own harnesses into reusable knowledge for supporting future tasks?}

To make this experience reusable, we distinguish the knowledge needed by the \Builder{} from task skills used by the \Target{}. Task skills describe how to perform a task~\citep{wei2026evoharness}; the Builder instead needs principles for designing the support that helps the Target perform it. We call these reusable support principles \MetaSkill[meta-skills]{}. Empirically, each meta-skill shuold specify \emph{when} support is needed, what capability or resource to \emph{provide}, and how the Target should \emph{use} it while retaining responsibility for judgment.

In this paper, we connect the learning and implementation of these principles through a construction--execution--reflection loop. Starting with an empty meta-skill bank, the \Builder{} constructs harnesses for development tasks, reviews the resulting Target execution records and scores, and adds or revises principles based on the observed evidence. After skill learning, the bank is frozen and guides fresh harness construction for each held-out task. Learning thus changes the Builder's external knowledge and the support it constructs, while both the Builder's and Target's model weights remain fixed.

We evaluate three Targets on Harness-Bench and NewtonBench under fixed Target execution budgets. The full-bank meta-skill Builder achieves a 65.31\% macro-average score, exceeding the no-skill Builder by 8.95 percentage points and full-bank independently learned Target skills by 10.93 points. It also outperforms direct delivery of the same bank to the Target in all six settings. These comparisons suggest that meta-skills can help the Builder become a more effective advisor: they guide support design, while Target skills guide task execution. In our setting, teaching the \Builder{} to translate experience into executable support can therefore be more beneficial than directly teaching the \Target{} additional task skills.

Because \MetaSkill[meta-skills]{} encode reusable principles, we further examine how they develop and whether they remain useful beyond the setting in which they were learned. Through analysis, we discover that repeated reflection can strengthen the guidance, and transfer studies suggest possible meta-skill reuse across Builders and Targets. The \Builder{} and \Target{} can also be instances of the same model: across three such settings, meta-skills improve scores by 18.71 points on average over no-skill construction. These results suggest that a model can improve its own execution by learning to build better support for itself, establishing harness design as a promising route to system-level self-improvement.

Looking ahead, AI4AI broadens the goal of agent learning: agents can learn both to solve problems and to create the conditions for others to succeed. Improving how Builders provide guidance and resources may be as consequential as improving how Targets use them. \textbf{Better agents may begin with better advisors.}

\begin{figure*}[!t]
\vspace{-0mm}
\centering
\includegraphics[width=0.96\textwidth]{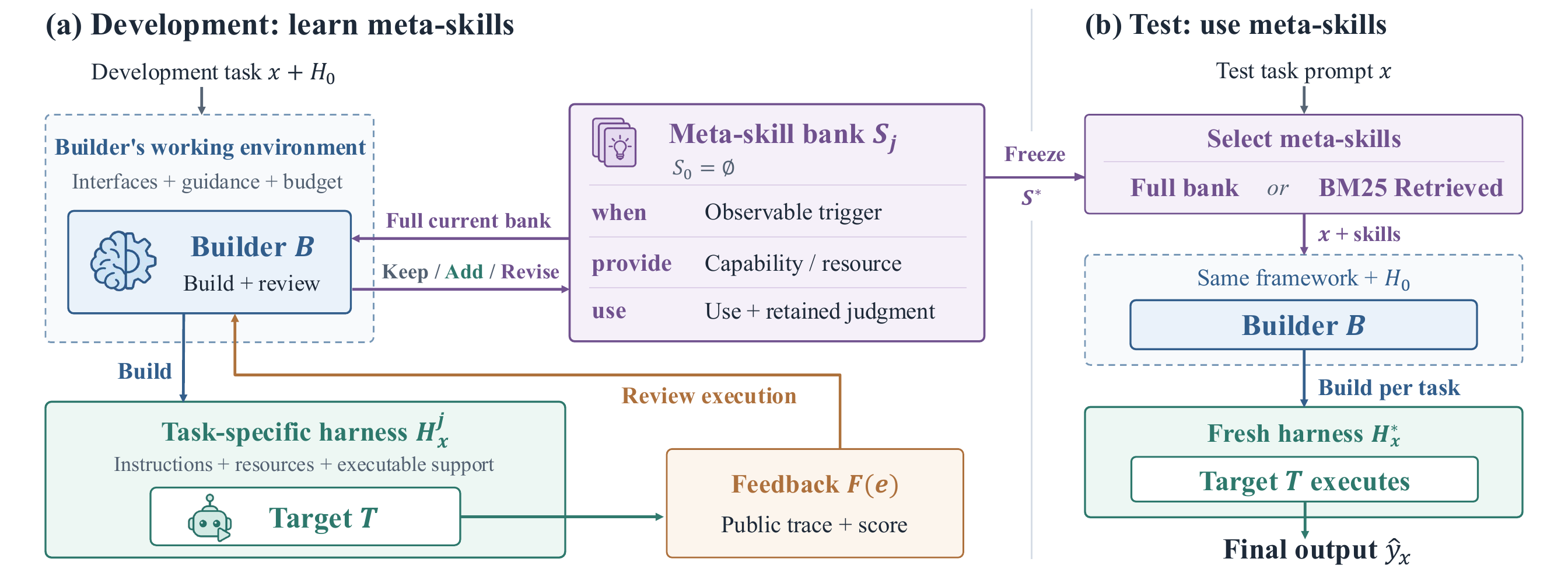}
\vspace{-0mm}
\caption{Overview of our framework. Rather than teaching the \Target{} directly, our framework equips the \Builder{} with \MetaSkill[meta-skills]{} learned from Target execution feedback. These reusable principles guide the Builder in constructing harnesses for new test tasks.}
\label{fig:method}
\vspace{-0mm}
\end{figure*}

%% file: sections/2_related_works.tex
\section{Related Work}

\paragraph{AI for AI in Agentic Systems.}
AI-for-AI uses AI systems to automate model development and agent design. For model development, AIDE uses an LLM agent to write and refine machine-learning pipelines, while MLE-Dojo provides environments for training and evaluating agents performing this engineering work~\citep{jiang2025aide,qiang2025mledojo}. For test-time agent design, the optimization target becomes the system surrounding a fixed model: Automated Design of Agentic Systems and AFlow both focus on searching agent programs and workflows~\citep{hu2025automated,zhang2025aflow}, while Darwin Godel Machine evolves its own agent code, and Meta-Harness searches harness implementations using execution feedback~\citep{zhang2026darwin,lee2026metaharness}. Within this direction, strong-to-weak harness construction uses a stronger Builder to supply executable support for a weaker Target~\citep{qian2026ai4ai}. We build on this setting by learning reusable construction principles from Target executions during offline skill learning, then freezing them to guide fresh harness construction for each test task, with both models' weights fixed.

\paragraph{Agent Skills and Meta-Skills.}
Agent skills preserve procedural knowledge for reuse: Voyager stores executable routines, while ExpeL and Agentic Context Engineering accumulate textual insights and contextual playbooks~\citep{wang2024voyager,zhao2024expel,zhang2026agentic}. Methods for improving these resources include SkillRL, which co-evolves skills and policies through reinforcement learning, and Evo-Harness, which compiles execution experience into transferable solver skills~\citep{xia2026skillrl,wei2026evoharness}. SkillsBench complements these methods by measuring skill utility with curated packages and deterministic verifiers~\citep{li2026skillsbench}. At the meta level, learned knowledge guides agent improvement itself: Meta Context Engineering learns skills for constructing context files and code, while MetaSkill-Evolve evolves policies for improving task skills~\citep{ye2026meta,wang2026metaskillevolve}. Within this direction, our \Builder{} learns those principles from its own harness outcomes and implements them as task-specific environments for a separate \Target{}, specifying when support is needed, what capabilities to supply, and the Target's responsibilities.

%% file: sections/3_preliminary.tex
% Empty, Combined with Method Sec.

%% file: sections/4_method.tex
\section{Problem Setting and Method}

\paragraph{Problem Formulation.}
Let $\B$ and $\T$ be fixed \Builder{} and \Target{} models. Each benchmark provides a baseline environment $H_0$ with native capabilities, an evaluator $r$, and a Target execution budget $C_x$. A harness policy $\mathcal H$ maps each public task input $x$ to an environment $H_x=\mathcal H(x)$ that augments $H_0$ with executable support. The objective is to maximize expected test performance:
\begin{equation}
J(\mathcal H)=\mathbb E_{x\sim\D_{\mathrm{test}},,\tau\sim\T(\cdot\mid x,H_x)}[r(x,\tau)],
\qquad \operatorname{cost}(\tau)\le C_x.
\label{eq:objective}
\end{equation}
The \Builder{} learns the policy through Target executions on development instances and is evaluated on held-out test instances. The budget $C_x$ covers only Target execution, excluding Builder computation for harness construction.

\paragraph{Method Overview.}
Instead of directly improving the \Target{}, we teach the \Builder{} to learn what support the Target needs and provide it through a harness. The Builder's experience accumulates in an external bank of \MetaSkill[meta-skills]{}, which records reusable principles for designing support across tasks. These principles guide the harness construction, while both the Builder and Target model weights remain fixed throughout learning.

\input{tables/harness}

\subsection{Learning Meta-Skills from Target Execution During Development}

\paragraph{Definition of Meta-Skill.}
A \MetaSkill{} $s=(\mathrm{when},\mathrm{provide},\mathrm{use})$ contains three fields: \emph{when} identifies observable conditions that call for support; \emph{provide} specifies the capability or resource the environment should supply; and \emph{use} explains how the \Target{} should employ that support and which judgments remain its responsibility.

\paragraph{Skill Learning Workflow.}
Skill learning begins with an empty skill bank $S_0$. For each development task input $x$, the \Builder{} constructs a separate harness using the neutral environment $H_0$, public component interfaces, and its complete current bank. The \Target{} executes within the harness, producing a public execution record and a benchmark-native development score. The same Builder then updates its skill bank by reviewing the current bank, its generated programs, and public execution feedback $F(e)$. This process will iterate until a fixed budget of development-set passes is reached, as formalized below:
\begin{equation}
\begin{aligned}
H_x^j = \B(H_0,x,S_j), \qquad e_x^j = \T(x;H_x^j), \qquad S_{j+1} = \operatorname{Revise}_{\B}\!\left(S_j,\{(H_x^j,F(e_x^j))\}_{x\in G_j}\right).
\end{aligned}
\label{eq:learn-build}
\end{equation}

\paragraph{Skill Bank Update.}
At the end of each skill-learning step, the \Builder{} reviews its generated harnesses and the Target's execution feedback to identify an observed error or recurring burden that better support could address, then compares the resulting lesson with the current bank. It \emph{revises} an existing meta-skill when the evidence corrects or strengthens its guidance, \emph{adds} a new meta-skill for a distinct reusable support need, or \emph{keeps} the bank unchanged when no update is justified. Each batch permits at most one addition or revision, which must cite supporting evidence from that batch to keep the learned guidance grounded in observed behavior.

\subsection{Building New Harness for Every Task at Test-Time}

\paragraph{Test-Time Skill Selection.}
After skill learning, we freeze the meta-skill bank and use it to guide test-time harness construction. For each test task, the Builder receives skills through one of two modes: \emph{full bank} places all meta-skills in its context; \emph{retrieval} uses a fixed BM25 retriever to select at most two skills with positive relevance scores against the task's initial public prompt. For each individual test task, the \Builder{} uses the supplied meta-skills to construct a new, task-specific harness in a fresh environment for the \Target{} to execute within:
\begin{equation}
S^*=S_J,\qquad H_x^*=\B(H_0,x,K(x,S^*)),\qquad
\hat y_x=\T(x;H_x^*),
\label{eq:deploy-support}
\end{equation}
where $K$ supplies either retrieved meta-skills or the full bank.

\paragraph{Harness Components.}
As summarized in \Cref{tab:harness-refinement-space}, the framework exposes seven optional harness component families: instructions, memory, context organization, composed tools, execution control, verification and recovery, and workspace preparation. The \Builder{} selects and implements these components, deciding what memory stores, when it is retrieved, and which tools and resources the Target receives. Within the resulting harness, the \Target{} reasons over observations, uses the available tools, and remains responsible for the final submission.

\paragraph{Builder's Working Environment.}
The \Target{} operates within a Builder-created harness, while the \Builder{} operates within a framework we design. Specifically, we provide shared interfaces, execution isolation, design guidance, and resource limits, with identical construction permissions and bounded interface-repair opportunities across conditions. These fixed priors underpin our workflow for automated meta-skill learning, harness construction, and Target execution. Our contribution thus shifts design effort to the Builder level, enabling it to \emph{turn execution experience into reusable support principles} and adapt their implementation to individual tasks.

\paragraph{Example.}
Consider an artifact edited after validation, making the earlier check potentially outdated. A \MetaSkill[meta-skill's]{} \emph{when} field identifies this condition; \emph{provide} requests version tracking and revalidation; and \emph{use} directs the \Target{} to inspect the current validation result before submitting, while retaining responsibility for content judgment and correction. The \Builder{} could then implement this principle with file-version memory and a validation tool. In Appendix we trace two actual test episodes from learned principles through Builder-written code and Target tool calls to final artifacts.

%% file: tables/harness.tex
% Requires: booktabs, array, graphicx
\begin{table*}[!t]
\vspace{-0mm}
\small
\centering
\caption{Overview of the harness refinement space, including neutral defaults in $H_0$, Builder-editable mechanisms, and illustrative examples. The underlying benchmark tools, evaluation budget, and scoring rules remain fixed.}
\vspace{-0mm}
\label{tab:harness-refinement-space}
\setlength{\tabcolsep}{2pt}
\renewcommand{\arraystretch}{1.15}
\resizebox{\textwidth}{!}{%
\begin{tabular}{@{}>{\raggedright\arraybackslash}p{2.05cm} *{7}{>{\raggedright\arraybackslash}p{2.15cm}}@{}}
\toprule
&
\textbf{Instructions} &
\textbf{Memory} &
\textbf{Context} &
\textbf{Composed tools} &
\textbf{Execution control} &
\textbf{Verification \& recovery} &
\textbf{Workspace} \\
\midrule
\textbf{Initial $H_0$} &
Generic prompt &
Empty store &
Default history &
Native tools &
Default loop &
Native rules &
Empty scratch \\
\midrule
\textbf{Builder $\B$ refines what?} &
Specific task guidance &
Record format; storage and retrieval rules &
History-selection rules &
Tool definitions; call sequences &
Execution phases; tool visibility &
Submission checks; failure handling &
Initial files; setup actions \\
\midrule
\textbf{Example Implement} &
``Draft before polishing'' &
Recent trial buffer &
Budgeted history window &
Cross-file auditor &
Hide failing tools &
Audit-gated submission &
Templates; validator scripts \\
\bottomrule
\end{tabular}%
}
\vspace{-0mm}
\end{table*}

%% file: sections/5_experiment.tex
\section{Experiments}
\label{sec:experiments}

\subsection{Experimental setup}

\begin{wraptable}{r}{0.42\linewidth}
\vspace{-4mm}
\centering
\small
\caption{Benchmark statistics and splits.}
\vspace{-2mm}
\label{tab:benchmark-splits}
\resizebox{\linewidth}{!}{
\begin{tabular}{lccc}
\toprule
\textbf{Benchmark} & \textbf{Tasks} & \textbf{Development} & \textbf{Test} \\
\midrule
Harness-Bench & 106 & 11 & 95 \\
NewtonBench & 324 & 32 & 292 \\
\bottomrule
\end{tabular}
}
\vspace{-2mm}
\end{wraptable}

\paragraph{Datasets.}
We evaluate the effect of test-time learned support on task performance using Harness-Bench and NewtonBench, which cover agent workflows and interactive scientific law discovery, respectively~\citep{yao2026harnessbench,zheng2025newtonbench}. For each benchmark, we reserve 10\% of tasks for learning and use the remainder exclusively for evaluation. See \Cref{tab:benchmark-splits} for details.

\paragraph{Models and Settings.}
We use GPT-5.6-Sol as the Builder model, and evaluate Gemini-3.6-Flash, Qwen3.8-Flash, and GPT-OSS-120B as Target models. For each Builder--Target--benchmark combination, the meta-skill bank starts empty and is updated over two development-set passes, with at most one evidence-grounded \emph{keep}, \emph{revise}, or \emph{add} update per task. The bank is then frozen for testing, while the Builder constructs a fresh harness for each test task.

\noindent All models use temperature zero and high reasoning effort (or the corresponding thinking mode), with a 16K output-token limit for harness generation and an 8K limit for skill update. For the benchmark budget, Harness-Bench allows 30 model turns, 30 tool calls, and 96K cumulative tokens per task; NewtonBench allows 12 turns, 10 tool calls, and 192K tokens.

\paragraph{Metrics.}
Harness-Bench reports the deterministic completion-oracle score over 95 test tasks, averaged and scaled to a percentage. NewtonBench reports symbolic-structure accuracy over 292 test tasks. Both use end-to-end evaluation, divided by the fixed number of test cases: valid native outcomes are retained, while audited harness-construction failures receive a score of zero because no executable harness is produced.
% Paired comparisons use 20,000 task-level bootstrap samples and unadjusted 95\% percentile intervals. We report the two native metrics separately; any mean difference across cells is descriptive rather than a new cross-benchmark metric.

\paragraph{Baselines.}
We compare five baselines. Learned guidance is supplied either in full or through BM25 top-2 retrieval.
\begin{itemize}[topsep=0pt, partopsep=0pt, leftmargin=*, itemsep=2pt]
\item \textbf{Native environment:} The Target solves tasks in the shared neutral environment, without learned guidance or Builder-generated support.
\item \textbf{No-skill Builder:} The Builder constructs harnesses with the same construction space and budget as our method, but with an empty skill bank.
\item \textbf{Direct Builder skills:} The Target receives our Builder's meta-skills directly as instructions, with no Builder-generated harness components.
\item \textbf{Independent structured Target skills:} An equally budgeted GPT-5.6-Sol learner iteratively extracts and refine structured problem-solving skills from the Target's own neutral-environment executions. The resulting bank is supplied directly to the Target.
\item \textbf{Mined free-form Target skills:} GPT-5.6-Sol consolidates the Target's neutral-environment development trajectories into free-form notes in one offline pass. These notes are supplied directly to the Target.
\end{itemize}
Our \textbf{Builder meta-skill} conditions instead provide the retrieved or full bank to the Builder, which compiles it into task-specific harness components such as instructions, memory, context, tools, controllers, verification, or workspace preparation. The full-bank Direct baseline therefore controls for semantic knowledge, while No-skill Builder controls for construction capability, serving as complementary ablations of our method. Please see \Cref{app:baseline-settings} for more setting details.

% \zhenhailong{this might need a bit more explanation; Not sure if this baseline is what I think of: I think the most important baseline is that: we use the same backend model of the Builder i.e., gpt-5.6-sol; check the Target task executions, and try to distill skills/knowledge that directly related to how to solve the task (instead of how to instruct a Target to solve the task); these skills will be available only to the Targets and can be for example retrieved and put directly in the context of a Target}

\subsection{Main results}
\input{tables/main}

We present the main results and baselines in \Cref{tab:test}, and highlight the following key findings.

\paragraph{Meta-skills are most useful when the teacher can enact them.}
Giving the same full meta-skill bank to the \Builder{} consistently outperforms giving it directly to the \Target{}, improving every model--benchmark pair by up to 25.43 points and 12.02 on average. This gap shows that the gains come not merely from exposing the Target to better knowledge, but from \emph{operationalizing} that knowledge before execution. The Builder can translate a declarative meta-skill into persistent state, executable tools, verification logic, or control decisions, reducing the burden on the Target to interpret and apply the advice correctly at inference time. \MetaSkill[Meta-skills]{} therefore function less as solver prompts and more as a compact language for \emph{provisioning} task-specific support.

\paragraph{Experience improves the Builder beyond construction capability.}
With construction capability held fixed, the full-bank Builder outperforms the no-skill Builder in all six settings, with an average gain of 8.95 points. These results suggest that experience adds value beyond the ability to construct scaffolds: it helps the \Builder{} decide \emph{which} support to build and \emph{how} to integrate it with Target behavior. The gains nevertheless vary across Targets and benchmarks, suggesting that accumulated guidance must be adapted to the Target and task rather than assumed uniformly useful.

\paragraph{Experience helps most when coordination is the main bottleneck.}
Relative to the no-skill Builder, gains on NewtonBench are positive across all three Targets and average 10.96 points, versus 6.95 points on Harness-Bench. NewtonBench requires agents to coordinate experimentation, reasoning, and valid symbolic submission, creating \emph{recurring coordination failures} that scaffolding can address. Harness-Bench spans more heterogeneous workflows, where effective support depends more on \emph{Target-specific planning and tool use}. This contrast suggests that accumulated experience is most useful when failure modes recur consistently across tasks and Targets.

\paragraph{Broad skill access usually outperforms sparse retrieval.}
The full skill bank outperforms top-2 retrieval in five of six settings, with an average gain of 7.19 points on NewtonBench. This pattern suggests that a compact meta-skill bank offers \emph{complementary procedures} whose combined value may be missed by lexical retrieval. Full-bank access allows the \Builder{} to consider these procedures together when designing support, making it a strong default at the evaluated scale. The exception, however, indicates that broader access is not uniformly beneficial and motivates retrieval methods that account for both skill complementarity and the Target's likely response to support.

%% file: tables/main.tex
\begin{table*}[t]
\vspace{-0mm}
\centering
\small
\setlength{\tabcolsep}{5pt}
\caption{Test performance (\%) on every task in the fixed test splits with GPT-5.6-Sol as Builder in all settings. Bold marks the best result in each model--benchmark column.}
\vspace{-0mm}
\resizebox{\linewidth}{!}{
\begin{tabular}{lccccccc}
\toprule
\multirow{2.5}{*}{\textbf{Method}} & \multicolumn{3}{c}{\textbf{Harness-Bench}} & \multicolumn{3}{c}{\textbf{NewtonBench}} & \multirow{2.5}{*}{\textbf{Macro Avg.}} \\
\cmidrule(lr){2-4}\cmidrule(lr){5-7}
& Gemini & Qwen & GPT-OSS & Gemini & Qwen & GPT-OSS & \\
\midrule
Native environment & 37.35 & 69.63 & 52.46 & 55.48 & 54.11 & 39.38 & 51.40 \\
Builder, no skills & 53.29 & 71.51 & 63.02 & 56.16 & 50.34 & 43.84 & 56.36 \\
Direct Builder skills, retrieved & 46.66 & 68.98 & 59.13 & 58.22 & 48.29 & 33.22 & 52.42 \\
Direct Builder skills, all & 42.38 & 68.70 & 58.34 & 58.56 & 55.82 & 35.96 & 53.29 \\
Independent Structured Target skills, retrieved & 41.95 & 72.51 & 45.21 & 67.81 & 53.42 & 36.30 & 52.87 \\
Independent Structured Target skills, all & 40.59 & 72.89 & 56.63 & 67.81 & 52.40 & 35.96 & 54.38 \\
Mined Free-form Target skills, retrieved & 39.61 & 63.51 & \textbf{71.12} & 46.92 & 54.45 & 33.90 & 51.59 \\
Mined Free-form Target skills, all & 40.70 & 66.10 & 66.01 & 41.44 & 54.11 & 39.04 & 51.23 \\
\midrule
\textbf{Builder meta-skills, retrieved} & 61.50 & \textbf{74.81} & 64.74 & 64.04 & 57.88 & 39.73 & 60.45 \\
\textbf{Builder meta-skills, all} & \textbf{67.81} & 72.64 & 68.22 & \textbf{68.84} & \textbf{63.70} & \textbf{50.68} & \textbf{65.31} \\
\bottomrule
\end{tabular}
}
\label{tab:test}
\vspace{-0mm}
\end{table*}

%% file: sections/6_analysis.tex
\section{Analysis}
\label{sec:analysis}

\begin{figure*}[t]
\vspace{-0mm}
\centering
\includegraphics[width=0.96\textwidth]{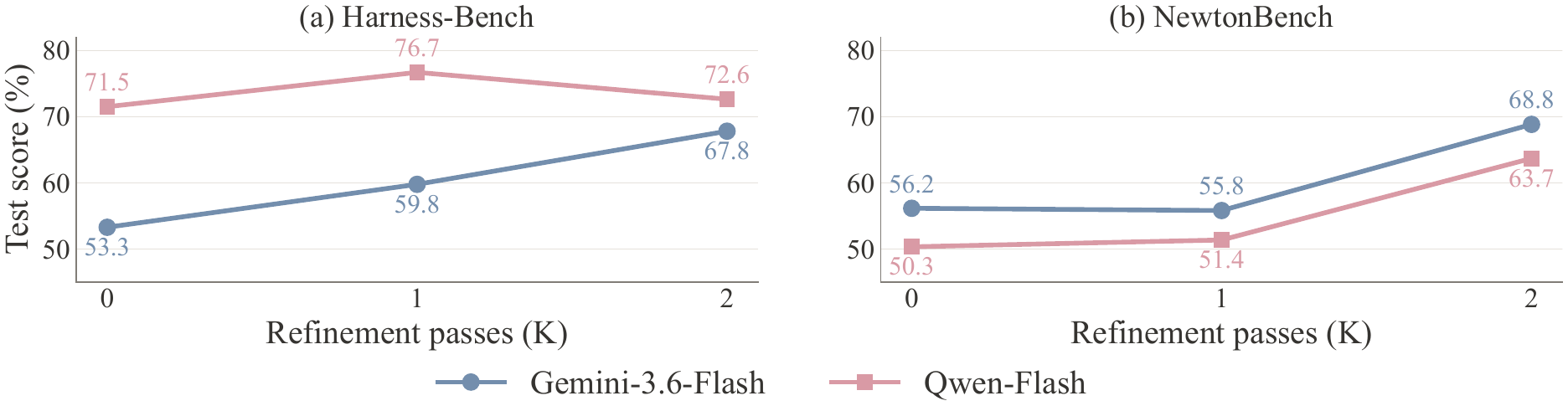}
\vspace{-0mm}
\caption{Test performance as Builder experience is refined. The two NewtonBench curves improve mainly after the second pass; Harness-Bench displays both steady improvement and late regression.}
\label{fig:refinement}
\vspace{-0mm}
\end{figure*}

\subsection{Effect of Meta-skill Refinement}
\label{sec:refinement}

\paragraph{Motivation and setting.}
Meta-skills develop through repeated Builder reflection, but additional updates need not improve teaching. To examine this progression, we freeze the skill bank after zero, one, or two complete development passes and evaluate each version on the full test split. \Cref{fig:refinement} reports results for both Gemini and Qwen Targets.

\paragraph{Useful teaching behavior can emerge late.}
On NewtonBench, the first pass changes each Target's score by less than 1.1 points, whereas the second adds 13.01 points for Gemini and 12.33 for Qwen. Across all settings, the average gain from zero to two passes is 10.42 points. Manual inspection suggests that early update gathers local observations, while later revision connects them into \emph{reusable interventions}. This delayed improvement is consistent with the value of iterative revision. The lower Mined Target skill score also motivates going beyond offline trajectory summarization, although that comparison changes both the learning procedure and skill recipient.

\paragraph{Refinement is not monotonic.}
On Harness-Bench, Gemini improves with each pass, but Qwen drops 4.06 points from its first-pass peak. Revisions can therefore sharpen useful guidance while also making it \emph{overly specific to recently observed evidence}. Together, these results suggest that effective learning requires both continued refinement and selective retention. A practical update rule should use development-only evidence to assess confidence in revisions and decide whether to continue, retain an earlier version, or roll back.

%%%%%%%%%%%%%%%%%%%%%%%%%%%%%%%%%%%%%
%%%%%%%%%%%%%%%%%%%%%%%%%%%%%%%%%%%%%

% \begin{figure*}[t]
% \centering
% \includegraphics[width=0.90\textwidth]{figures/same_model_self_evolution.pdf}
% \caption{Same-model self-evolution study. Each group compares no-skill construction, independent Target skills, and Builder meta-skills; labels on the meta-skill bars give the absolute score and improvement over no-skill construction.}
% \label{fig:self-evolution}
% \end{figure*}

\subsection{Self-Improvement through Harness Design}
\label{sec:self-evolution}

\paragraph{Motivation and setting.} Meta-skills need not rely on a stronger external teacher: the same model can serve as both \Builder{} and \Target{}. We evaluate three settings using Gemini-3.6-Flash and Gemini-3.1-Pro, with construction, reflection, and execution performed by the same model within each case. This setting tests whether a model can use past execution experience to improve its own future performance through \emph{learned support design}.

\begin{figure*}[t]
\centering
\vspace{-0mm}
\begin{minipage}[t]{0.445\textwidth}
    \vspace{0pt}
    \centering
    \includegraphics[width=\linewidth]{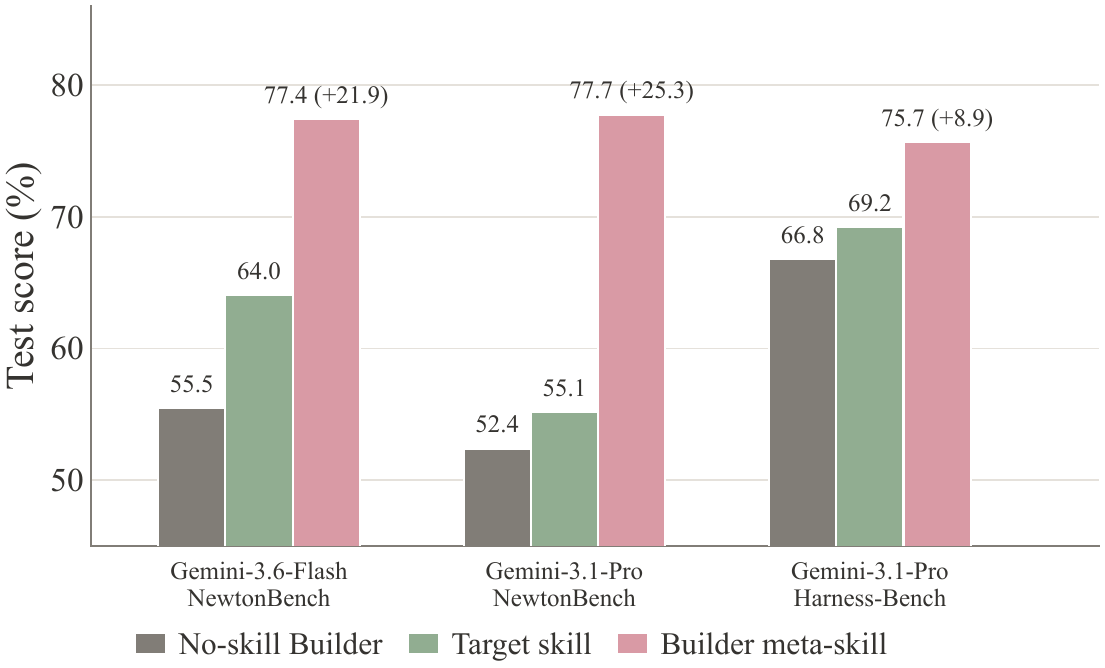}
    \par\smallskip
    \textbf{(a) Same-model self-evolution}
\end{minipage}\hfill
\begin{minipage}[t]{0.542\textwidth}
    \vspace{0pt}
    \centering
    \includegraphics[width=\linewidth]{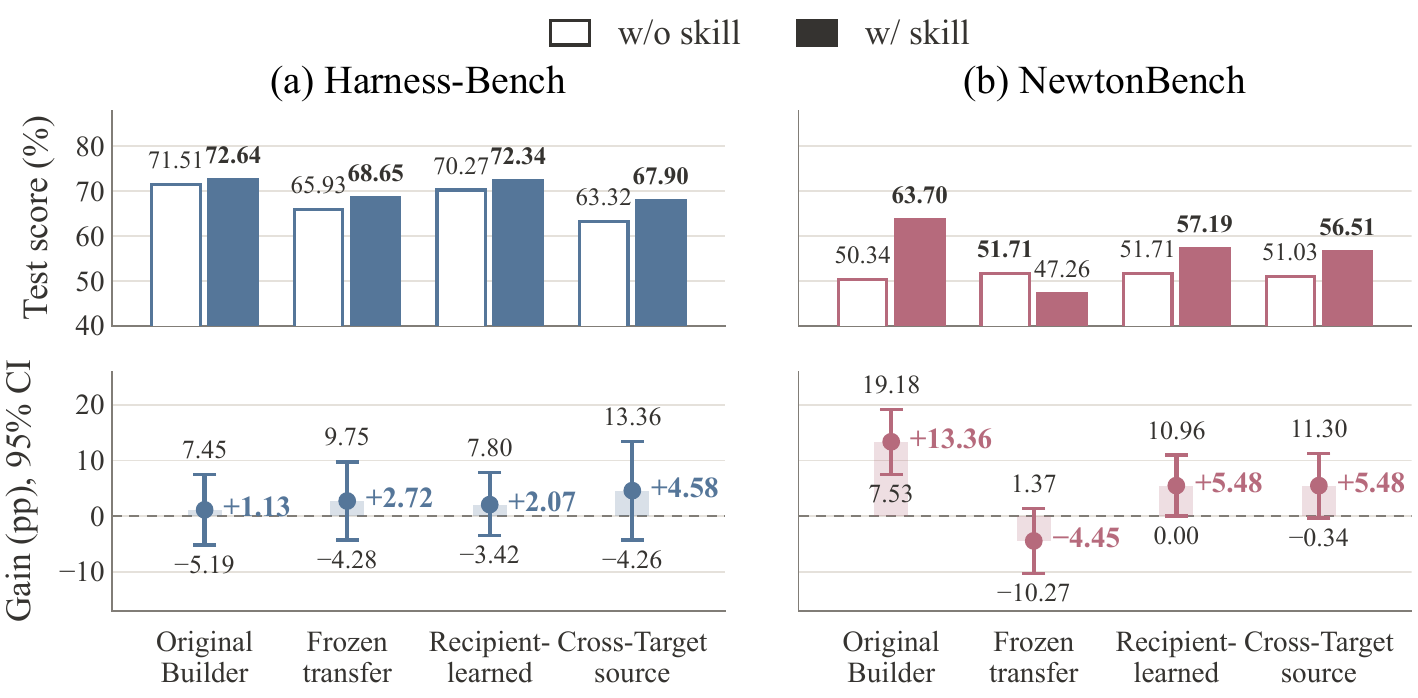}
    \par\smallskip
    \textbf{(b) Cross-Builder transfer}
\end{minipage}
\vspace{0mm}
\caption{
\textbf{(a)} Same-model self-evolution compares no-skill construction, independent Target skills, and Builder meta-skills. Meta-skill bar labels show absolute scores and gains over no-skill construction.
\textbf{(b)} Cross-Builder transfer on Harness-Bench and NewtonBench. Top: test scores without (outlined bars) and with (filled bars) Builder meta-skills. Bottom: paired gains in percentage points with 95\% task-bootstrap confidence intervals; dashed lines indicate zero gain.
}
\label{fig:self-evolution-transfer}
\vspace{-0mm}
\end{figure*}

\paragraph{Support design is a distinct target for self-improvement.}
Across the three settings in \Cref{fig:self-evolution-transfer}(a), Builder meta-skills yield average gains of 18.71 points over no-skill construction and 14.14 points over skills delivered directly to the Target. These results identify an additional axis of optimization: improving how a model equips itself, without weight updates or a stronger teacher. This mechanism could complement the Target's own skill learning: as the Target's capabilities improve, the Builder could adapt its support to the Target's changing needs, while execution feedback informs further support refinement. Self-evolution could therefore involve coordinated improvement in both task-solving capabilities and the environments that support them.

%%%%%%%%%%%%%%%%%%%%%%%%%%%%%%%%%%%%%
%%%%%%%%%%%%%%%%%%%%%%%%%%%%%%%%%%%%%

\subsection{Meta-Skill Transfer across Builders and Targets}
\label{sec:transfer}

\paragraph{Motivation and setting.} \MetaSkill[Meta-skills]{} specify support principles while leaving implementation to the \Builder{}. We test whether a learned bank remains useful when another Builder translates it into harnesses. All evaluations use Qwen-Flash as Target on the complete test splits of Harness-Bench and NewtonBench, comparing four settings:
\begin{itemize}[topsep=-3pt, partopsep=0pt, leftmargin=*, itemsep=2pt]
\item \textbf{Original Builder reuse:} GPT-5.6-Sol learns from Qwen-Flash's development executions and uses its own bank to construct test harnesses.
\item \textbf{Transfer across Builders:} Gemini-3.1-Pro constructs harnesses using the frozen Sol-to-Qwen bank, changing the Builder while holding the bank and Target fixed.
\item \textbf{Recipient Builder learning:} Gemini-3.1-Pro learns its own bank from Qwen-Flash's development executions and uses it to construct test harnesses.
\item \textbf{Transfer across Builders and Targets:} Gemini-3.1-Pro receives a frozen bank learned by Sol from Gemini-3.6-Flash's executions and constructs harnesses for Qwen-Flash, changing both the Builder and the Target relative to skill acquisition.
\end{itemize}
Banks are learned on each benchmark's development split and frozen before testing. Each receiving Builder uses the full bank to construct a fresh harness per test task. \Cref{fig:self-evolution-transfer}(b) reports scores and paired differences against contemporaneous no-skill runs of the same Builder.

\paragraph{Meta-skills support cross-Builder reuse, with benefits dependent on implementation.}
On NewtonBench, the Sol-to-Qwen bank yields a 13.36-point gain with Sol but a 4.45-point decline when transferred to Gemini-Pro; on Harness-Bench, the same transfer instead yields a 2.72-point gain. These results support the feasibility of cross-Builder reuse, although all confidence intervals include zero, leaving the performance benefits uncertain. The differing outcomes suggest that transfer depends partly on \emph{how the receiving \Builder{} translates shared lessons into a harness}: even the skill bank is fixed, different Builders may implement the same guidance differently. Refining the resulting harness through execution feedback may therefore help realize the value of transferred meta-skills.

\paragraph{Recipient-side learning yields stronger observed performance.}
Gemini-Pro's own bank outperforms the imported Sol-to-Qwen bank by 3.69 points on Harness-Bench and 9.93 points on NewtonBench, with positive gains over its no-skill controls on both benchmarks. This pattern suggests that \emph{learning from the recipient's own harness executions still better align meta-skills with its implementation choices}. Recipient-side learning may therefore complement cross-Builder reuse: imported skills provide reusable guidance, while feedback from the recipient's harnesses offers a basis for refining that guidance.

\paragraph{Useful support principles may transfer across both Builders and Targets.}
Transferring Sol's Gemini-derived bank to Gemini-Pro building harnesses for Qwen yields gains of 4.58 points on Harness-Bench and 5.48 points on NewtonBench. On NewtonBench, the transferred bank scores only 0.68 points below Gemini-Pro's own Qwen-derived bank and even outperforms the imported Sol-to-Qwen bank, despite the latter matching the receiving Target. This pattern suggests that \emph{the relevance of learned support principles may matter more than an exact source-Target match}: guidance learned for one Builder–Target pair can remain useful when another \Builder{} implements it for a different \Target{}.

%%%%%%%%%%%%%%%%%%%%%%%%%%%%%%%%%%%%%
%%%%%%%%%%%%%%%%%%%%%%%%%%%%%%%%%%%%%

\begin{table*}[!t] %{r}{0.65\linewidth}
\centering
\small
\setlength{\tabcolsep}{5pt}
\renewcommand{\arraystretch}{1.15}
\vspace{-0mm}
\caption{Harness component ablations on NewtonBench. We report test scores (\%), changes (ablation minus full) using the main-table full-bank execution as the reference, and unadjusted 95\% paired task-bootstrap confidence intervals for these changes. Intervals containing zero indicate uncertainty in the direction of the average effect.}
\vspace{-0mm}
\resizebox{\linewidth}{!}{
\begin{tabular}{lc|cc|cc}
\toprule
\textbf{Target model} & \textbf{Full Harness} & \textbf{w/o Controller} & \textbf{95\% CI} & \textbf{w/o Memory + Context} & \textbf{95\% CI} \\
\midrule
Gemini-3.6-Flash & 68.84 & $55.48^{\text{-13.36}}$ & $[-18.49,-8.22]$ & $66.78^{\text{-2.05}}$ & $[-7.53,3.08]$ \\
Qwen-Flash & 63.70 & $58.90^{\text{-4.79}}$ & $[-10.27,0.68]$ & $59.59^{\text{-4.11}}$ & $[-9.25,1.03]$ \\
GPT-OSS-120B & 50.68 & $49.66^{\text{-1.03}}$ & $[-6.51,4.11]$ & $54.11^{\text{+3.42}}$ & $[-2.05,8.90]$ \\
\bottomrule
\end{tabular}
}
\vspace{-0mm}
\label{tab:component-ablation}
\end{table*}

\subsection{Harness Component Ablations}
\label{sec:components}

\paragraph{Motivation and setting.} Meta-skills affect execution through the harnesses the \Builder{} creates. Using NewtonBench as a case study, we freeze the constructed harnesses and replay every test task after removing either the execution controller or memory plus context. \Cref{tab:component-ablation} reports the average score difference between complete and ablated harnesses, with negative values favoring retention. The accompanying \emph{confidence intervals} quantify uncertainty in this average effect across test tasks.

\paragraph{Controllers generally improve harness performance across Targets.}
Removing the controller lowers scores for all three Targets: by 13.36 points for Gemini, 4.79 for Qwen, and 1.03 for GPT-OSS. These consistent observed drops support the controller's role in improving execution, although only Gemini's confidence interval excludes zero. The size of the benefit varies with both \emph{Target behavior and controller design}: each ablation isolates the controller within a frozen harness, but its functions may differ across Targets, including experiment sequencing, state tracking, or submission management. Effective harness design therefore uses external coordination where it helps, tailoring the controller's functions to each Target's execution needs.

\paragraph{Memory and context show promise as targeted support.}
Retaining memory plus context yields small positive estimated effects for Gemini and Qwen and a negative effect for GPT-OSS, although all three confidence intervals contain zero. These estimates suggest potential benefits for some Targets, but do not establish a reliable aggregate effect. Additional state can preserve useful evidence, while its value depends on keeping that evidence relevant and avoiding redundant or outdated information. Meta-skills therefore should support \emph{selective state exposure} by guiding what to retain and when to surface it for the \Target{}.

%%%%%%%%%%%%%%%%%%%%%%%%%%%%%%%%%%%%%
%%%%%%%%%%%%%%%%%%%%%%%%%%%%%%%%%%%%%

\subsection{Outcome Transitions and Error Analysis}
\label{sec:errors}

\begin{figure*}[t]
\vspace{0mm}
\centering
\begin{minipage}[t]{0.28\textwidth}
    \vspace{0pt}
    \centering
    \includegraphics[width=\linewidth]{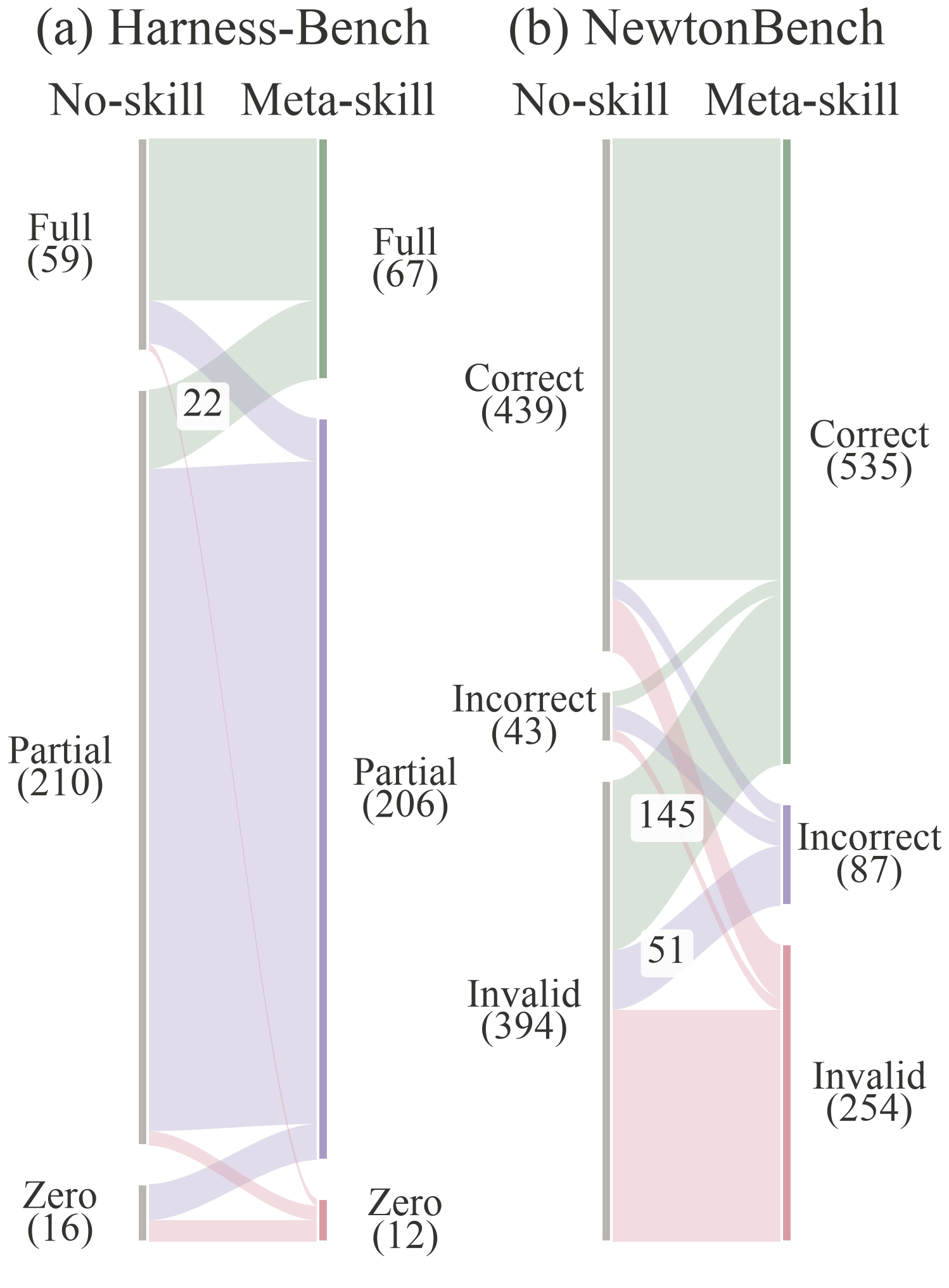}
    \par\smallskip
    \textbf{(a) Outcome transitions}
\end{minipage}\hfill
\begin{minipage}[t]{0.712\textwidth}
    \vspace{0pt}
    \centering
    \includegraphics[width=\linewidth]{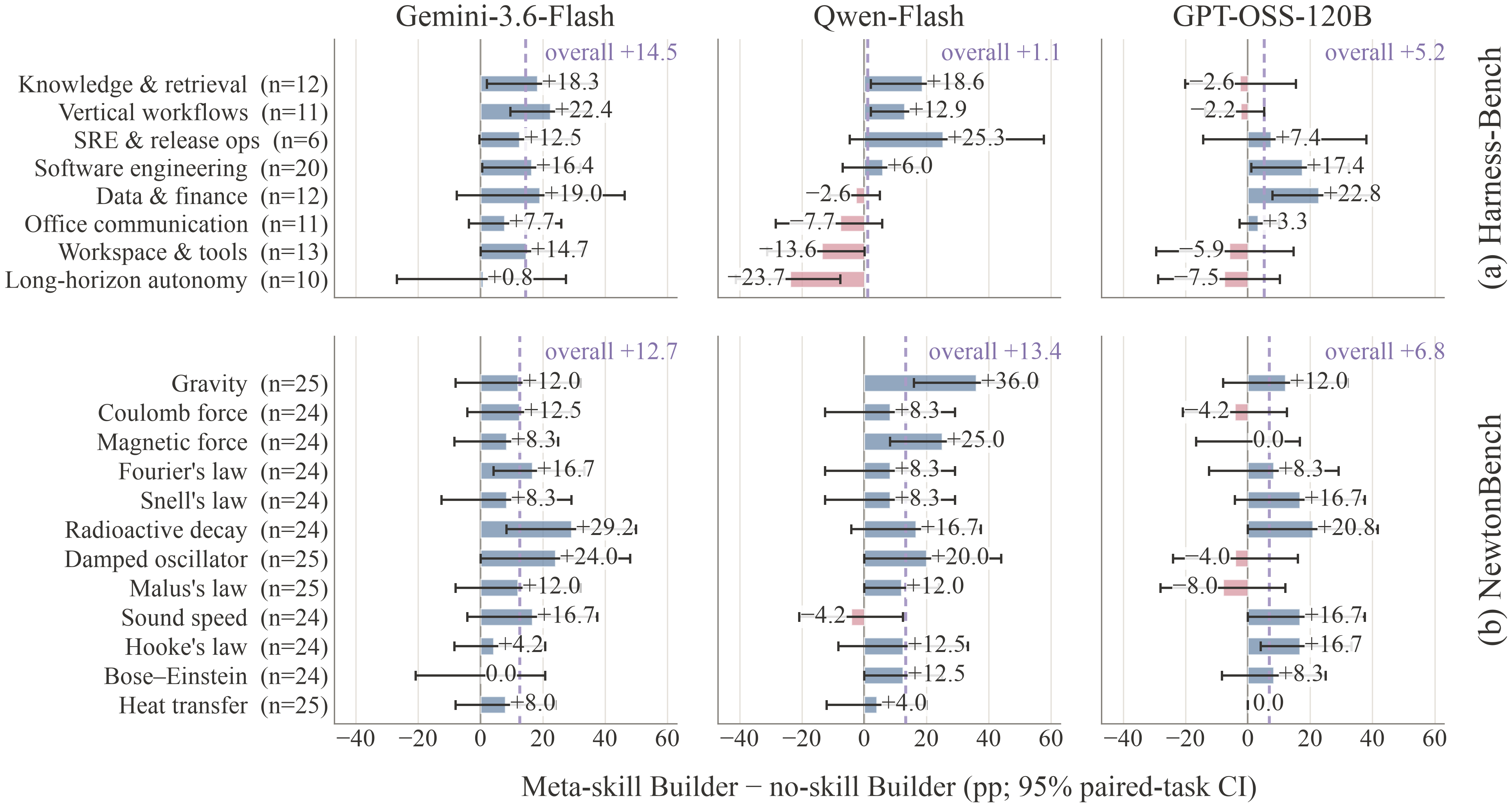}
    \par\smallskip
    \textbf{(b) Category-level effects}
\end{minipage}
\caption{
\textbf{(a)} Outcome transitions from no-skill to full-bank meta-skill Builders, aggregated over three Targets. Band widths represent paired-task counts; in-band numbers highlight the largest cross-outcome flows. ``Zero'' includes zero scores and construction failures; ``Invalid'' includes missing or unparsable submissions and construction failures.
\textbf{(b)} Category-level effects across all six settings. Rows represent Harness-Bench task categories and NewtonBench physical mechanisms; columns represent Targets. Bars show score differences between full-bank and no-skill Builders with unadjusted 95\% task-bootstrap confidence intervals. Purple dashed lines indicate setting's overall effect.
}
\label{fig:transitions-categories}
\vspace{-0mm}
\end{figure*}

\paragraph{Motivation and setting.} Aggregate scores can conceal both changes in execution outcomes and offsetting gains and regressions across task families. We pair the full-bank meta-skill and no-skill Builders on all model--task pairs across the six main-study settings. In \Cref{fig:transitions-categories}, panel (a) aggregates outcome transitions across the three Targets within each benchmark, showing which outcomes are rescued or degraded, while panel (b) breaks down paired score differences by Harness-Bench task category and NewtonBench physical mechanism for each Target, revealing where improvements concentrate and where failures persist.

\paragraph{Meta-skills help close the execution gap.}
On NewtonBench, cases with no valid submission fall from 394 to 254 (35.5\%), while correct discoveries rise from 439 to 535 (21.9\%). Transitions include 145 recoveries from no valid submission to a correct law, versus 46 regressions. This suggests more reliable completion of the scientific loop, creating opportunities for correct discovery. Valid-but-incorrect submissions also rise from 43 to 87, showing that completion enables but does not guarantee correctness. On Harness-Bench, full successes increase from 59 to 67, largely through transitions from partial to full success. This suggests that meta-skills help agents satisfy remaining requirements and \emph{turn partial progress into complete task fulfillment}.

\paragraph{Meta-skill benefits recur across Targets, with task-dependent variation.}
All three Targets improve overall on NewtonBench by 6.85--13.36 points, with positive category-level point estimates for gravity, radioactive decay, and Hooke's law across models. Harness-Bench shows gains for every Target on SRE/release tasks, with mixed effects elsewhere. GPT-OSS improves by 5.20 points overall but declines by 2.23 points on vertical workflows, illustrating that a positive aggregate gain can coexist with category-specific regressions. These recurring gains suggest shared support needs across models, while the remaining variation points to the importance of Target--task compatibility. This further motivates \emph{reusing support principles while adapting their implementation} to each Target's capabilities and the task's demands.

%% file: sections/7_conclusion.tex
\section{Discussion and Conclusion}

\paragraph{Message 1: Meta-skills guide resource allocation.}
With semantic knowledge held fixed, Builder enactment gains 12.02 points over direct delivery on average. This suggests that experience becomes useful through decisions about which burdens to externalize as state, tools, or control. Meta-skills guide these decisions, helping \Builder[Builders]{} allocate support so that \Target[Targets]{} can devote their limited interaction budgets to the remaining judgment.

\paragraph{Message 2: Transfer connects reusable principles with adaptive implementation.}
Observed gains across Builders and Targets suggest that support principles can generalize across models, while variable effects indicate that their implementation matters. Stronger observed performance from recipient-learned banks further motivates adapting transferred guidance. Future systems could preserve a skill's support intent while using recipient-side feedback to refine what information to store, which tools to provide, and when to intervene.

\paragraph{Message 3: Environment design offers a path to self-improvement.}
Same-model studies show that a model can improve its own test-time execution by learning to construct better harnesses, while its weights remain fixed. Our evidence spans one main Builder, two benchmarks, and one adopted execution per task and condition. Establishing broader applicability requires more diverse Builders and tasks, and evaluation of performance gains relative to construction cost, which we leave to future works.

\paragraph{Conclusion.}
In this work, we introduced \MetaSkill{} that enable a \Builder{} to learn reusable support principles from the execution outcomes of its own harnesses and implement them for unseen tasks. With both models' weights fixed, full-bank meta-skills improve macro-average performance across Harness-Bench and NewtonBench by 8.95 percentage points over no-skill construction and 12.02 points over direct delivery of the same bank to the Target. These comparisons highlight the value of translating experience into executable support: a lesson can guide the construction of tools, state, and control mechanisms that help the Target apply its capabilities during execution. Outcome analyses connect these gains to more reliable task completion, while component and transfer results point to the importance of adapting reusable principles to each model and task. Same-model experiments further demonstrate that a model can self-improve by learning to build better support for itself. Together, these findings identify support design as a complementary axis of agent improvement.

\section*{Ethics Statement}
This work studies automated harness design using existing agent-workflow and simulated scientific-discovery benchmarks. Experiments use isolated execution environments and bounded resources, with development and test tasks separated and meta-skill banks frozen before evaluation to preserve evaluation integrity. Because generated tools and controllers can enable unintended or harmful actions, applications beyond these benchmarks should incorporate restricted permissions, validation of generated components, and human oversight appropriate to the application.

% \section*{Reproducibility Statement}
% Section~3 describes the meta-skill representation, learning and update procedures, harness design space, and test-time construction protocol. Section~4.1 and Table~2 document the benchmark split sizes, model configurations, learning and execution budgets, baseline definitions, and scoring conventions, including the treatment of construction failures. Sections~5.1--5.4 describe the refinement, same-model, transfer, and component-ablation protocols, while Section~5.5 explains the paired outcome and category-level analyses. Together, these descriptions support reproduction of the learning, construction, and evaluation pipeline and comparison under consistent experimental conditions.

%% file: sections/Appendix.tex
\appendix
\clearpage
\section*{Appendix}

\section{Significance, Scope, and Clarifications}

\subsection{Why Builder Learning Matters}
We discuss the significance of Builder learning and highlight the core contributions of this study.

\textbf{Learning to Design Useful Support.}
We study how a Builder can turn experience into reusable principles for supporting (or harnessing) a Target. These meta-skills specify when support is needed, what to provide, and how the Target should use it. They guide fresh harness construction on subsequent tasks, allowing the Builder to carry forward design knowledge while adapting its implementation to each task. This makes support design itself an object of learning.

\textbf{Turning Experience into Executable Support.}
In the main study, giving the full skill bank to the Builder yields higher scores than giving the same bank directly to the Target in all six settings. This highlights the value of translating guidance into tools, state, and execution control that the Target can use. The same-model experiments further illustrate a route to system-level self-improvement: a model can use execution experience to learn how to equip itself more effectively, with its weights unchanged.

\subsection{Clarifications of Results}
\label{app:scope-clarifications}
We explain the rationale for our experimental settings and clarify the conclusions supported by our results.

\textbf{Contribution within test-time AI4AI.}
We study how experience improves a Builder's ability to support a Target while both models' weights remain fixed. Closely related work learns solver guidance, as in Evo-Harness~\citep{wei2026evoharness}; searches executable harness implementations, as in Meta-Harness~\citep{lee2026metaharness}; or evolves strategies for engineering context files and code, as in Meta Context Engineering~\citep{ye2026meta}. Our focus is reusable knowledge for support design: the Builder extracts principles from its own harness outcomes, specifying when support is needed, what to provide, and how the Target should use it. These principles subsequently guide fresh construction for held-out tasks. Our contribution combines this learning and deployment protocol with controlled evidence on the value of accumulated Builder experience and its implementation as task-specific support.

\textbf{Shared framework and interpretation of the comparisons.}
All Builder conditions share component interfaces, design guidance, construction permissions, and bounded repair opportunities. The no-skill Builder assesses the additional value of learned guidance under the same construction allowance. Direct-bank baselines preserve the skill content and assess the benefit of having a Builder implement it before execution, while independently learned Target skills and mined notes provide alternative routes for applying experience to task solving. Full-bank macro-average gains of 8.95 points over no-skill construction and 12.02 points over direct delivery support the value of learning and implementing support principles within this framework. These comparisons fix Target execution limits; learning and construction fall outside those limits, with construction allowances matched across Builder conditions. Frozen-harness ablations examine component contributions within the resulting implementations, and case studies illustrate how principles become executable support. Their Target-dependent results, including the clear controller benefit for Gemini-3.6-Flash, support adapting available components to the recipient's needs.

\textbf{Evaluation protocol and reproducibility.}
Skill acquisition uses development episodes, and the bank is frozen before testing. Each test harness is constructed from the public task input and the supplied bank, without access to test execution feedback or reference answers. Appendices~B and~C document the core settings: the partitioning procedure, information access, model choices, resource limits, baseline configurations, failure accounting, and paired task-bootstrap procedures. All declared test tasks remain in the evaluation denominator, and reported intervals characterize uncertainty across paired test instances. Upon paper acceptance, we will release the full code and all prompts for skill learning, harness construction, baseline runs, and evaluation, together with the experimental configurations.

\textbf{Generalization to new tasks within each benchmark.}
Banks learned from approximately 10\% of each benchmark guide construction on the remaining held-out instances, with full-bank improvements over no-skill construction in all six settings. This establishes the usefulness of accumulated experience for new tasks within known categories and mechanisms, covering both workflow execution and scientific discovery. Cross-dataset reuse is a natural extension beyond this evaluation scope and is left to future work. A promising hypothesis is that skill granularity is central to broader reuse: principles should specify recurring support needs precisely enough to guide construction while leaving their implementation adaptable. Separating support principles from executable realizations provides a concrete basis for studying this balance.

\textbf{Transfer as reuse across Builder--Target configurations.}
Our transfer study examines reuse of a frozen bank when the Builder or the Builder--Target pair changes. Effects are measured against concurrent no-skill runs of the receiving Builder and interpreted through paired confidence intervals. Cross-Builder-and-Target transfer yields positive point estimates on both benchmarks, illustrating promising reuse across selected model combinations. The broader results establish the feasibility of sharing principles while showing that their realized utility depends on the receiving configuration, with performance benefits qualified by the reported uncertainty. Recipient-side learning offers a complementary route for aligning support principles with a Builder's own implementations.

\textbf{A simple foundation for meta-skill evolution.}
We adopt a simple initial evolution mechanism to make the contribution of experience directly assessable. Starting from an empty bank, the Builder constructs harnesses, reviews Target execution, and makes one evidence-grounded decision per development task: keep, revise, or add. Two development passes produce the bank used for evaluation. Held-out gains demonstrate that useful support principles can emerge through this lightweight procedure, while same-model experiments extend the result to self-improvement without a stronger external teacher. The refinement curves show that update effects depend on the Target and task, motivating future advances in selective revision and retention on top of this working foundation.

\subsection{Scope and Future Directions}
\label{app:scope-future}
We define the scope of the present study and identify promising directions for future research that build on its findings.

\textbf{Support design as the object of learning.}
This work studies learning that changes the knowledge and execution environments surrounding fixed models. The reported improvements therefore concern the resulting Builder--Target system, including system-level self-improvement when the same model serves both roles. The meta-skill representation and component interfaces provide a concrete instantiation of this learning problem within a shared design space. Our contribution establishes the usefulness of this route under the evaluated conditions and provides a reference point for studying alternative support representations and construction policies. A complementary direction is to combine Builder learning with improvements to the Target itself, allowing task-solving capabilities and the support that makes them effective to develop together.

\textbf{Empirical scope and explanatory claims.}
We evaluate end-to-end task performance, encompassing both benchmark correctness and the execution processes needed to achieve it. Preserving evidence, validating artifacts, and reaching a valid submission are substantive parts of this objective. The empirical effects are conditional on the evaluated task inventories, models, learned banks, and recorded executions. Component analyses examine contributions within the generated harnesses, while case studies illustrate concrete ways in which support changes execution. These findings provide an empirical basis for further work on stability and mechanism: independent learning runs, repeated executions, and alternative task partitions can characterize variability across experience histories, and targeted interventions on individual principles and component interactions can refine the causal account of how support helps.

\textbf{Generality through shared support needs.}
The main generalization claim concerns new instances within known workflow categories and physical mechanisms; the model-transfer studies provide exploratory evidence for selected receiving configurations. Extending reuse across datasets, unseen task families, and different tool interfaces is a promising next step. One hypothesis is that the scope of a principle depends strongly on the granularity at which it describes a support need. For example, preserving evidence after a tool failure may be useful across domains even when the appropriate storage and recovery mechanisms differ. Studying this separation between a reusable need and its local implementation could clarify when principles transfer directly, when recipient-side adaptation helps, and how to learn banks that serve broader collections of tasks.

\textbf{Evolution over longer learning histories.}
The current two-pass development procedure followed by frozen-bank deployment demonstrates that a bounded amount of experience can improve subsequent harness construction. Extending this setting to longer task streams would allow study of how useful principles are retained, revised, and combined as experience accumulates. The observed variation between refinement passes motivates development-based assessment of candidate updates and selective retention of successful guidance. Larger banks also create opportunities for adaptive retrieval and skill composition, extending the current comparison of full-bank access and BM25 top-2 retrieval. These directions would develop the present initial mechanism into a learning process suited to longer histories and changing support needs.

\textbf{Performance objectives and deployment conditions.}
The experiments study performance under fixed Target interaction limits, with Builder learning and construction providing support outside that execution budget. This setting makes the value of prepared support assessable for a given Target operating under the same interaction constraints. The supported performance claim uses this budget definition. For deployment, a complementary objective is to allocate a joint compute or latency budget across learning, construction, and execution according to the workload. Future work can study how reuse of learned banks, caching of common support components, and task-adaptive construction effort affect this allocation, connecting the demonstrated performance benefits to different deployment requirements.

%%%%%%%%%%%%%%%%%%%%%%%%%%%%%%%%%%%%%%%%

\section{Reproducibility Details}
\label{app:implementation}

\subsection{Data partitions and information access}
For each benchmark, we reserve a development set of approximately 10\% of the instances for meta-skill learning and use the remainder for testing. We determine the development-set size by rounding ten percent of the total to the nearest integer, with ties rounded up, and allocate proportional integer quotas across the public Harness-Bench categories and NewtonBench domains. This partitioning evaluates generalization to held-out instances within known categories and mechanisms. We use no separate validation partition, and the learned skill bank is frozen before testing, with no revisions based on test outcomes.

Each meta-skill contains \emph{when}, \emph{provide}, and \emph{use} fields, with a combined limit of 192 tokens. During learning, the Builder receives the entire current bank and updates it using evidence from learning episodes. The actual submission, or its absence, serves as the primary evidence, interpreted alongside the public task requirements, complete input--observation pairs, public errors, final artifacts, and support delivered by the harness. Each update produces a single decision: keep the bank unchanged, revise an existing skill, or add a new skill. A revision must identify the existing skill and the current evidence motivating the change; superseded versions remain archived. Neither private reference answers nor researcher-written diagnoses are provided during this process.

At deployment, BM25 queries the skill bank using only the initial public task prompt and retrieves at most two skills with positive match scores. Harness construction receives the public initial task, neutral component interfaces, and the assigned semantic skill pack. The Builder has no access to test trajectories, scores, reference answers, or evaluator internals, so construction depends entirely on the information available before Target execution.

\subsection{Models and resource limits}
We use GPT-5.6-Sol as the Builder for both harness construction and meta-skill updates. The Target models are Gemini-3.6-Flash, Qwen3.8-Flash, and GPT-OSS-120B. All requests specify a temperature of zero and request high reasoning effort.

Harness construction allows up to 16K output tokens, with at most two repair attempts for candidates that violate the component interfaces. Each skill-update call allows up to 8K output tokens and returns the single keep, revise, or add decision described above. Target execution follows the benchmark-specific resource allowances listed below; these allowances are identical across experiment settings.

\input{tables/hyperparameters}

\subsection{Outcome and uncertainty accounting}
\label{app:failure-accounting}

All reported aggregates use a fixed denominator containing every declared test task. When a run yields a valid native benchmark score, we retain that score, including zero. When construction or execution fails to yield a valid result, the task contributes zero to the system-level aggregate. This rule covers missing source bindings, infrastructure failures, and unresolved technical outcomes. The Builder and Target may attempt repairs within their permitted limits, but a run that remains unsuccessful or exceeds its total budget is still counted as zero. Thus, the absence of a native score never removes a task from the reported denominator.

For paired comparisons, we resample test tasks 20K times while preserving the pairing between conditions. We report the 2.5th and 97.5th percentiles as unadjusted 95\% intervals. These intervals describe variation across tasks in the fixed test inventory, rather than variability across repeated stochastic model generations.

Component ablations operate directly on frozen harnesses: we remove one component family without rebuilding the remaining support. Each paired contrast therefore measures the effect of that intervention on the recorded harness, conditional on a single official execution per task. For transfer experiments, paired comparisons use only the concurrent no-skill control from the same campaign. Differences across campaigns are reported only descriptively.

%%%%%%%%%%%%%%%%%%%%%%%%%%%%%%%%%%%%%%%%

\section{Baseline configurations and interpretation}
\label{app:baseline-settings}

All conditions share the same development/test splits and per-task Target execution limits. We use GPT-5.6-Sol for both harness construction and skill extraction, with Gemini-3.6-Flash, Qwen3.8-Flash, and GPT-OSS-120B as Targets. We keep all the learned banks and notes frozen before testing.

\textbf{Retrieved versus full-bank access.}
These variants use the same learned content and differ only in how much of it is supplied at deployment. Full-bank access provides every learned skill, whereas retrieval uses the initial public task prompt as a BM25 query ($k_1=1.5$, $b=0.75$) and selects at most two positive-scoring matches. If no skill receives a positive score, the resulting pack is empty. For free-form trajectory-mined notes, retrieval operates on natural paragraphs, while full-text access provides the complete notes. Holding the learned content fixed allows this comparison to assess selective versus comprehensive delivery without changing the learning procedure.

\textbf{Native environment.}
The Target executes directly in the shared neutral environment $H_0$, using native tools, default conversation history, and scratch storage. It receives no learned guidance or Builder-generated components, although it may create its own working aids through the available tools. This baseline represents ordinary task solving in the Target's native environment and provides the reference point for evaluating additional support. Gains from our method over this setting therefore reflect the combined contribution of harness construction and learned meta-skills.

\textbf{No-skill Builder.}
To separate the contribution of construction from that of learned experience, GPT-5.6-Sol constructs a fresh harness for each public task using an empty skill pack. The Builder otherwise retains the same seven component families, construction instructions, and information about the Target as our method, together with the same allowance of 16K output tokens and at most two interface-repair attempts. This baseline preserves the Builder's ability to design support while removing access to learned meta-skills. For Gemini on Harness-Bench, performance increases from 37.35 in the native environment to 53.29 with no-skill construction, and then to 67.81 with our method. The first improvement captures the benefit of construction alone; the second captures the additional benefit of learned experience under the same construction interface.

\textbf{Direct Builder skills.}
This baseline supplies the Target with the same meta-skill pack that the Builder receives in the corresponding variant of our method. The meta-skills' \emph{when}, \emph{provide}, and \emph{use} fields are appended to the Target's instructions, and the Target is explicitly allowed to implement this guidance using its native tools and execution budget. No Builder-generated callbacks or resources are installed. The comparison thus holds access to learned knowledge fixed while changing whether a separate Builder implements that knowledge as harness support. Our full-bank method achieves higher scores in all six original settings, including 67.81 versus 42.38 for Gemini on Harness-Bench. These results support the benefit of having the Builder translate meta-skill guidance into concrete support before Target execution.

\textbf{Independent Structured Target skills.}
This baseline learns skills specifically for the Target's own problem-solving, using a separate GPT-5.6-Sol learner with no access to our Builder's programs, trajectories, or meta-skill bank. Learning proceeds over two passes through the development set: the Target executes tasks in the neutral environment, and its current skill bank is supplied to subsequent executions as updates accumulate. Each update allows up to 8K output tokens and at most one skill addition or revision, with each skill limited to 192 tokens. The resulting bank is frozen and delivered directly to the Target at test time. Here, ``independent'' refers to the separately collected experience and separately learned bank; the extraction model remains the same. This baseline compares learning guidance for the Target's task-solving procedures with learning principles for the Builder's support design. Direct Target skill learning is competitive in several settings: Qwen scores 72.89 on Harness-Bench compared with our method's 74.81, while Gemini scores 67.81 on NewtonBench compared with 68.84.

\textbf{Mined Free-Form Target skills.}
This baseline also uses GPT-5.6-Sol to extract problem-solving advice from independent Target experience, but replaces iterative structured updates with a single offline consolidation step. The Target first completes one pass through the development set, after which the learner consolidates all resulting trajectories into a note of at most 3K tokens. The learner freely chooses the content, format, and organization of this note, without prescribed skill fields or explicit add, revise, and keep operations. At test time, the frozen notes are supplied directly to the Target, with no separate harness construction or Builder-generated components. Results vary across Target--benchmark settings. On Harness-Bench, GPT-OSS scores 66.01 with free-form notes, compared with 56.63 with Independent Structured Target skills. On NewtonBench, Gemini scores 41.44 with free-form notes, compared with 67.81 with structured Target skills and 68.84 with our method. These comparisons show that the relative effectiveness of static free-form advice and iteratively learned structured skills depends on the evaluation settings and models.

%%%%%%%%%%%%%%%%%%%%%%%%%%%%%%%%%%%%%%%%

\input{sections/case_studies}

\section{AI Use Statement}
In this work, LLMs are used strictly for research support rather than as sources of substantive content. Their use falls into three categories: (i) providing tested results on our AI4AI at test-time framework, and (ii) assisting with language refinement during paper writing. For writing support, we used ChatGPT solely to polish text (improving coherence and grammar) while all ideas, logic, results, and technical contributions originate from the authors. To safeguard rigor, we have carefully reviewed all LLM-refined texts to confirm that no hallucinated content was introduced and that the original arguments, findings, and perspectives were faithfully preserved.

%% file: tables/hyperparameters.tex
\begin{table}[htbp]
\centering
\small
\caption{Per-task Target limits. Each model call permits at most 16,384 output tokens; NewtonBench additionally limits one experiment call to 20 queried items. Limits are identical across conditions within a setting.}
\begin{tabular}{lrrrr}
\toprule
\textbf{Benchmark} & \textbf{Model turns} & \textbf{Tool calls} & \textbf{Tokens} & \textbf{Wall time} \\
\midrule
Harness-Bench & 30 & 30 & 96,000 & 1,200 s \\
NewtonBench & 12 & 10 & 192,000 & 3,600 s \\
\bottomrule
\end{tabular}
\label{tab:hyperparameters}
\end{table}

%% file: sections/case_studies.tex
\section{Case Studies: From Meta-Skills to Executable Support}
\label{app:cases}

We present two successful full-bank test episodes with GPT-5.6-Sol as \Builder{} and Gemini-3.6-Flash as \Target{}. They illustrate complementary forms of support design: constructing a task-specific executable checker and coordinating an existing scientific interface through memory and control. Each case traces how a learned meta-skill informs the generated harness and how that support is used during \Target{} execution. These selected cases provide qualitative insight into the resulting workflows, without estimating typical success rates or isolating the causal contributions of individual components.

\subsection{Making an artifact contract executable}
\label{app:case-artifacts}

\begin{casestudy}{Case 1: Builder connects artifact validation to submission}
\textbf{Task and support problem.}
A metric-migration task in Harness-Bench asks \Target{} to compare old and new metric definitions, query results, and dashboard snapshots under a migration policy. It requires four deliverables: a metric comparison, a regression ledger, a summary, and a caveat report. The challenge combines interpreting changes with keeping these deliverables complete and consistent. For example, a higher activation rate may be justified by a revised denominator, whereas a retention dashboard that still uses the old denominator violates the migration policy. Distinguishing these cases requires semantic judgment; ensuring that the resulting classifications agree across artifacts requires systematic bookkeeping.

\medskip
\textbf{From a learned principle to executable support.}
The frozen bank provides a checkpointing meta-skill that calls for tracking output updates, validation status, and remaining budget, with a prompt finish after successful validation. It leaves evidence interpretation, drafting, and repair to \Target{}. \Builder{} implements this principle through task-specific instructions, an executable auditor, a controller, and a submission gate. The auditor is generated during harness construction, while \Target{} subsequently authors all substantive deliverables. This division assigns mechanical verification to the harness and preserves \Target{}'s responsibility for the analysis.

\medskip
\textbf{The auditor makes consistency requirements explicit.}
The generated auditor checks that the metric comparison covers every metric appearing in either query export exactly once, that reported values match the sources, that the regression ledger has the required coverage, and that the summary's category counts agree with the comparison. It also verifies the presence of all required artifacts, their schemas, and numerical formatting. These checks specialize the public output requirements to the task's actual inputs. For example, once \Target{} labels a metric as an unexpected regression, the auditor can check whether the ledger and summary consistently reflect that decision. Determining whether the label is justified by the migration policy remains \Target{}'s responsibility.

\medskip
\textbf{Validation governs the submission workflow.}
The controller tracks writes to the four deliverables and prompts \Target{} to invoke the auditor once all four have been observed. A clean audit enables submission and prompts completion unless \Target{} has identified a concrete semantic defect. Subsequent edits through the tracked file-writing interface invalidate the audit result and require another check. The submission gate enforces this dependency by withholding completion until the controller records a clean audit. Together, these components connect artifact creation, validation, and submission through an explicit readiness condition.

\medskip
\textbf{Observed execution and outcome.}
In the recorded episode, \Target{} reads the migration evidence, writes the four deliverables, and invokes the auditor, which reports zero errors and zero warnings. The resulting analysis covers five metrics: two expected definition changes, one unexpected regression, one with no material change, and one requiring review. The regression ledger identifies the retention dashboard's continued use of the old cohort denominator, while the missing post-migration gross-margin value is left for review. \Target{} then submits successfully, achieving a native score of 1.0 with 14 model calls and 15 counted tool calls. No repair iteration is needed. On the same task, no-skill construction and Direct-all produce no output artifacts and each score 0.0435, terminating because of token-budget exhaustion and model failure, respectively.

\medskip
\textbf{Insight: support design separates judgment from bookkeeping.}
The reusable knowledge lies in recognizing which obligations can be made executable. \Builder{} translates the checkpointing principle into checks for metric coverage, source-value fidelity, and agreement across deliverables. These obligations are straightforward to verify deterministically but compete for attention with interpreting migration evidence. By assigning them to generated code, the harness lets \Target{} concentrate on substantive decisions while making their consistent expression across artifacts mechanically checkable.

\medskip
\textbf{Insight: validation becomes actionable when it controls completion.}
The auditor supplies a capability, while the controller and submission gate determine when to use it and how its result changes the next action. A clean check establishes a concrete stopping condition, and an observed edit reopens validation. This episode demonstrates the successful draft--audit--submit path: the constructed workflow supports delivery of a complete, consistent answer within the execution budget. The comparison illustrates the potential value of this integrated support, although a single episode does not isolate the contribution of each component.
\end{casestudy}

\subsection{Preserving scientific progress after computational failure}
\label{app:case-science}

\begin{casestudy}{Case 2: Builder preserves evidence and bounds failed computation}
\textbf{Task and support problem.}
A NewtonBench task exposes a noise-free apparatus controlled by displacement $x$ and mass $m$, which returns velocity after exponential energy loss. The requested quantity is the pre-loss elastic potential $U(x)$, so the directly calculable kinetic energy $E_k=mv^2/2$ must be corrected for the loss. Within ten tool calls, \Target{} must design informative experiments, distinguish the measured observable from the requested quantity, infer the underlying law, and submit an executable function. The support problem combines a conceptual challenge---recovering latent energy from observations---with a procedural one: preserving enough evidence and budget to finish when computational analysis fails.

\medskip
\textbf{From learned principles to coordinated support.}
Three learned meta-skills guide the construction: retain raw evidence while bounding analysis, reserve budget for submission, and distinguish the requested quantity from measured observables. \Builder{} implements these principles through instructions, memory, a controller, and recovery and submission checks around the benchmark's existing experimental and computational interfaces. The instructions direct \Target{} to infer the energy-loss scale jointly with the displacement law and recommend duplicate inputs, cross-mass probes, and fresh measurements. The scientific hypothesis and its coefficients remain for \Target{} to infer.

\medskip
\textbf{Memory preserves evidence independently of analysis.}
The memory module stores recent experiment batches together with their inputs, separately from the latest computation result. A failed fitting attempt therefore leaves the measurements available for subsequent reasoning. Retrieval supplies the latest analysis result and up to four recent experiment batches to later model requests. In the recorded run, both earlier experiment batches are delivered again immediately after the second computation failure, making the accumulated evidence available at the point of recovery.

\medskip
\textbf{Control bounds retries and preserves a path to submission.}
The controller counts failed computation calls and, after two failures, removes computation from the available tool menu while keeping experimentation and submission available. Recovery feedback directs \Target{} to continue from retained measurements. When three or fewer tool or model calls remain, a finalization reminder asks \Target{} to spend at most one further action on a concrete contradiction or fresh validation before preparing the submission. A separate submission check verifies the required response format and function signature. These mechanisms make recovery responsive to observed failures and remaining budget.

\medskip
\textbf{Observed analysis and recovery.}
\Target{} first collects a broad experiment batch containing duplicate inputs and cross-mass probes, followed by a second batch focused on small displacements. Two successful computation calls derive kinetic energy and compare candidate loss corrections. For example, at $x=0.01$, the observed kinetic energy is approximately $4.58571$, while the candidate corrections $E_k e^{100x}$ and $E_k e^{50x}$ yield approximately $12.46524$ and $7.56055$, respectively. These calculations explicitly connect the measured observable to hypotheses about pre-loss energy. The next two fitting attempts fail: one uses unsupported syntax, and the other exceeds the computational instruction limit. The controller then removes computation from the tool menu, and \Target{} collects ten additional measurements at displacements between $0.0001$ and $0.001$. With three tool calls remaining, the finalization reminder activates, and \Target{} proceeds to submission.

\medskip
\textbf{Scientific result.}
\Target{} submits the following law:
\[
\widehat{U}(x)=\max\{0,\;1241.9012\,x+462.282\,x^2\}.
\]
The native evaluator accepts the submission as equivalent to the reference law, reporting symbolic accuracy 1 and RMSLE 0. The run uses eight model calls and eight counted tool calls. On the same task, no-skill construction produces a valid but incorrect law, with symbolic accuracy 0 and RMSLE 0.092, while Direct-all ends with a model failure and no submission. The successful trajectory therefore includes an observed recovery from failed analysis to further measurement and completion.

\medskip
\textbf{Insight: recovery connects evidence retention with action control.}
The initial guidance to keep computation bounded does not prevent two fitting failures. The constructed support nevertheless preserves a productive continuation: memory retains the measurements, the controller recognizes repeated failure, and the available actions change accordingly. \Target{} can return to experimentation without losing its earlier evidence or spending the remaining budget on further computation retries. This case illustrates how a reusable recovery principle becomes an executable policy that responds to the actual trajectory.

\medskip
\textbf{Insight: support components should match the bottleneck.}
The conceptual and procedural challenges receive different forms of support. Instructions clarify the distinction between observed and latent energy; memory maintains evidence continuity; and control limits unsuccessful analysis while reserving an opportunity to finish. Their coordination allows \Builder{} to improve the use of an existing scientific interface. \Target{} retains responsibility for choosing experiments and inferring the law, while the harness structures how evidence, failures, and remaining resources inform its next action.
\end{casestudy}